\documentclass[letterpaper]{article} 
\usepackage{aaai23}  
\usepackage{times}  
\usepackage{helvet}  
\usepackage{courier}  
\usepackage[hyphens]{url}  
\usepackage{graphicx} 
\usepackage{natbib}  
\usepackage{caption} 
\usepackage{times}
\usepackage{arydshln}
\usepackage{wrapfig, lipsum, booktabs}
\usepackage{float}
\usepackage{latexsym}
\usepackage{tabularx}
\usepackage[ruled, vlined, linesnumbered]{algorithm2e}
\usepackage{setspace}
\usepackage{subfigure}   
\usepackage{multirow}
\usepackage{graphicx}
\usepackage{amsfonts}
\usepackage{amsmath}
\usepackage[utf8]{inputenc}

\usepackage{cleveref}
\crefname{section}{§}{§§}
\Crefname{section}{§}{§§}

\usepackage{pifont}
\usepackage{xcolor} 
\newcommand{\xmark}{\ding{55}}%
\newcommand{\cmark}{\ding{51}}%
\definecolor{Purple}{RGB}{192, 172, 217}
\definecolor{Gray}{RGB}{153, 153, 153}
\definecolor{Blue}{RGB}{0, 113, 188}
\definecolor{SeaGreen}{RGB}{67, 205, 128}
\definecolor{GoldNrod}{RGB}{218, 165, 32}

\usepackage{newfloat}
\usepackage{listings}
\DeclareCaptionStyle{ruled}{labelfont=normalfont,labelsep=colon,strut=off} 
\floatstyle{ruled}
\newfloat{listing}{tb}{lst}{}
\floatname{listing}{Listing}
\title{Language Model Pre-training on True Negatives}
\author{
   Zhuosheng Zhang\textsuperscript{\rm 1,2},
	Hai Zhao\textsuperscript{\rm 1,2,\thanks{Corresponding author.  This paper was partially supported by Key Projects of National Natural Science Foundation of China (U1836222 and 61733011).}},
	Masao Utiyama\textsuperscript{\rm 3},
	Eiichiro Sumita\textsuperscript{\rm 3}\\
}
\affiliations{
    \textsuperscript{\rm 1}Department of Computer Science and Engineering, Shanghai Jiao Tong University\\
	\textsuperscript{\rm 2}Key Laboratory of Shanghai Education Commission for Intelligent Interaction\\
	and Cognitive Engineering, Shanghai Jiao Tong University, Shanghai, China\\
	\textsuperscript{\rm 3}National Institute of Information and Communications Technology (NICT), Kyoto, Japan\\
	\texttt{zhangzs@sjtu.edu.cn, zhaohai@cs.sjtu.edu.cn, \{mutiyama,eiichiro.sumita\}@nict.go.jp}
}

\usepackage{bibentry}

\begin{document}

\maketitle

\begin{abstract}
Discriminative pre-trained language models (PLMs) learn to predict original texts from intentionally corrupted ones. Taking the former text as positive and the latter as negative samples, the PLM can be trained effectively for contextualized representation. However, the training of such a type of PLMs highly relies on the quality of the automatically constructed samples. Existing PLMs simply treat all corrupted texts as equal negative without any examination, which actually lets the resulting model inevitably suffer from the false negative issue where training is carried out on pseudo-negative data and leads to less efficiency and less robustness in the resulting PLMs. In this work, on the basis of defining the false negative issue in discriminative PLMs that has been ignored for a long time, we design enhanced pre-training methods to counteract false negative predictions and encourage pre-training language models on true negatives by correcting the harmful gradient updates subject to false negative predictions. Experimental results on GLUE and SQuAD benchmarks show that our counter-false-negative pre-training methods indeed bring about better performance together with stronger robustness.
\end{abstract}

\section{Introduction}
Large-scale pre-trained language (PLM) models are playing an important role in a wide variety of NLP tasks with their impressive empirical performance \citep{radford2018improving,peters2018deep,devlin-etal-2019-bert,yang2019xlnet,lan2019albert,clark2019electra}. So far, there comes two major categories of PLMs with regards to the output style, the generative like GPT \citep{radford2018improving}, which employ a decoder for learning to predict a full sequence, and the discriminative like BERT style of PLMs which learn to reconstruct the original uncorrupted text from the intentionally corrupted ones \citep{raffel2020exploring,lewis2020bart}. In this work, we focus on the latter category of PLMs, typically with denoising objectives (also known as masked language
modeling, MLM) \citep{liu2019roberta,joshi-etal-2020-spanbert,sun2019ernie}. In a denoising objective, a certain percentage of tokens in the input sentence are masked out, and the model should predict those corrupted tokens during the pre-training \citep{peters2018deep,sun2019ernie,levine2021pmimasking,li-zhao-2021-pre}.\footnote{There are different classification standards for PLMs, i.e., output style and model architecture. For simplicity, our taxonomy follows \citet{wang2021can}, which is based on the output style. Note that PLMs can also be classified into three types based on the model architecture: encoder-only, decoder-only and encoder-decoder.} 




Although existing studies have made progress in designing effective masking strategies \citep{sun2019ernie,joshi-etal-2020-spanbert,levine2021pmimasking} and auxiliary objectives \citep{lan2019albert,wang2019structbert} for language model pre-training, there is still a lack of attention on the quality of training data. Discriminative PLM can be regarded as a kind of auto denoising encoder on automatically corrupted texts. Thus, it is critical to ensure the auto-constructed data is true enough. Intuitively, a discriminative PLM learns to distinguish two types of samples, positive (already existing original ones) and negative (the corrupted ones from the auto constructing). Taking MLM as an example, a proportion of tokens in sentences are corrupted, e.g., replaced with mask symbols, which would affect the sentence structures, leading to the loss of semantics and increasing the uncertainty of predictions. In extreme cases, such corrupted texts may be linguistically correct. However, the current PLMs simply consider all corrupted texts as negative samples, so that the resulting PLM has to be trained on such pseudo-negative data with less efficiency and less robustness -- suffers from the wasted training time on meaningless data and the trained PLM may be vulnerable to adversarial attacks like diversity distraction and synonym substitution \citep{wang2021textflint}.

\begin{table*}
 \setlength{\tabcolsep}{10pt}
 \setlength{\belowcaptionskip}{3pt}
\centering
\begin{tabular}{lllcc}
\toprule
\textbf{Example} & \textbf{Ground-truth} & \textbf{Prediction} & \textbf{MLM} & \textbf{Correction} \\
\midrule
i am trying to copy \colorbox[RGB]{221,226,237}{{[MASK]}} onto my ipod  good & \colorbox[RGB]{253,232,216}{{you}} & \colorbox[RGB]{195,227,233}{{happy}} & \xmark & - \\
\midrule
an adaptive immune system whose \colorbox[RGB]{221,226,237}{{[MASK]}} function ...    & \colorbox[RGB]{253,232,216}{{primary}} & \colorbox[RGB]{248,204,202}{{main}} & \xmark & \cmark\\
\bottomrule
\end{tabular}

\caption{{Examples of true negative (the first line) and false negative (the second line). The standard MLM will treat all the predictions as incorrect ones. However, the last false negative predictions can be corrected to ensure more accurate pre-training. More examples are in Figure \ref{fig:case}.}}
\label{tab:mis_example}

\end{table*}

\begin{figure}
	\centering
	\includegraphics[width=0.46\textwidth]{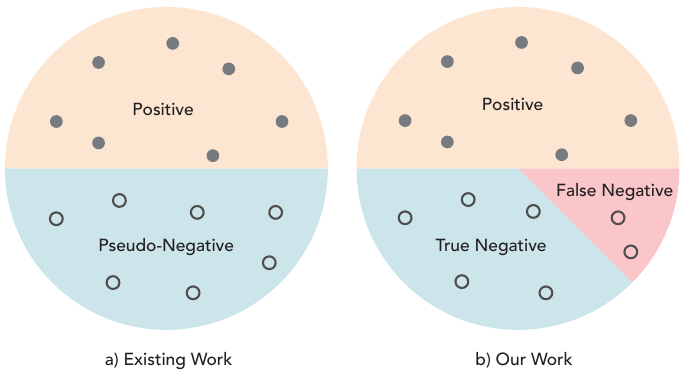}
	\caption{Overview of our study. Existing PLMs were trained by distinguishing positive from pseudo-negative data. In contrast, our work aims to encourage pre-training language models on true negatives by detecting and counteracting false negative predictions. \label{fig:over}} 
\end{figure}

For each training instance, MLM only calculates label-wise matching between the prediction and the gold tokens in the training process, thus inevitably suffering from the issue of false negatives where the prediction is meaningful but regarded as wrong cases, as examples shown in Table \ref{tab:mis_example}. We obverse that such cases appear in more than 7\% of the training examples (more details in Section \ref{sec:stat}). The issue is also observed in sequence generation tasks, which is tied to the standard training criterion of maximum likelihood estimation (MLE) that treats all incorrect predictions as being equally incorrect \citep{wieting2019beyond,li2020data}. Instead of measuring negative diversity via diversity scores between the different incorrect model outputs, our method is dedicated to mediating the training process by detecting the alternative predictions as opposed to the gold one, to steer model training on true negatives, which benefits the resulting language modeling in general. The comparison with existing work is illustrated in Figure \ref{fig:over}.

Though the false negatives may potentially hurt the pre-training in both efficiency and robustness to a great extent, it is surprising that this problem is kept out of the research scope of PLMs until this work to our best knowledge. To address the issue of misconceived false negative predictions and encourage pre-training language models on true negatives or more true negatives, we present an enhanced pre-training approach to counteract misconceived negatives. In detail, we investigate two enhanced pre-training objectives: 1) hard correction to shield the gradient propagation of the false negative samples to avoid training with false negative predictions; 2) soft regularization by minimizing the semantic distances between the prediction and the original one to smooth the rough cross-entropy. Experimental results on widely-used down-streaming benchmark tasks, including GLUE \citep{wang-etal-2018-glue} and SQuAD \citep{Rajpurkar2016SQuAD}, show that our approach boosts the baseline performance by a large margin, which verifies the effectiveness of our proposed methods and the importance of training on true negatives. Case studies show that our method keeps simplicity and also improves robustness.

\section{Preliminaries: The \textit{False Negative} Issue}\label{sec:stat}
\paragraph{Definition} Our concerned false negatives in MLM are the reasonable predictions but discriminated as wrong predictions because such predictions do not match the single gold token for each training case. For example, many tokens are reasonable but written in different forms or are synonyms of the expected gold token. 

\paragraph{Severity} For simplicity, we focus on the subset of false negatives {from WordNet \citep{miller1995wordnet}} -- the predictions which are the synonyms of the ground-truth tokens. To have an intuition about the severity of false negative predictions during pre-training, we collect the statistics from two perspectives: 1) prediction-level: the proportion of corrected predictions when they mismatch the gold labels; 2) iteration-level: the proportion of iterations (sequences) when the correction happens.\footnote{The rationale is that training on false negatives tends to learn incorrect semantics of the whole sequence.} We use the wikitext-2-raw-v1 corpus \citep{merity2016pointer} for validation. We use the pre-trained checkpoints of the BERT-base and BERT-large models described in Section \ref{sec:setup} for the analysis.\footnote{{The MLM process is the same as our experiments on BERT models in the subsequent sections.}} 

\begin{table}
\centering
 \setlength{\tabcolsep}{8.8pt}
     
    \begin{tabular}{l c c c c c c c c c c}
    \toprule
    \multicolumn{3}{c}{\textbf{\emph{Base Model}}} & \multicolumn{3}{c}{\textbf{\emph{Large Model}}} \\
    Check. & Iter. & Pred. &Check. & Iter. & Pred. \\
    \midrule
    6.25 &	6.90	&1.31	 & 6.25& 	7.46&	1.50 \\
    12.5&	6.96	&1.34	&12.5&	7.58&	1.55 \\
    25.0&	6.97&	1.36	&25.0	&7.31	&1.49 \\
    50.0&	7.05&	1.36	&50.0	&7.46	&1.56 \\
    80.0&	7.06&	1.40	&80.0	&7.38	&1.57 \\
    100.0&	7.07&	1.41	&100.0&	7.44	&1.60 \\
    \bottomrule
    \end{tabular}
    \caption{Statistics (\%) of the hard corrections under $\rm{base}$ and $\rm{large}$ settings on the wikitext-2-raw-v1 corpus. Checkpoint means the checkpoint saved at the specific training steps (\%).}
    \label{tab:stat-hc}
    
\end{table}

%
    

According to Table \ref{tab:stat-hc}, we observe that the ratio of detected false negatives is around 6.0\%-7.0\% in iteration-level and 1.0\%-2.0\% in token-level.\footnote{It is hard to collect the statistics of false negatives automatically. For simplicity, we only calculate the subset related to synonyms. Therefore, the issue is expected to occur more frequently than counted.} As training goes on, the correction ratio increases, indicating that our method gradually plays a more important role as the training proceeds, which supports our hypothesis. 


\paragraph{Influence} Pre-training on false negatives would possibly bring harm in terms of training efficiency, model effectiveness, and robustness against adversarial attacks (detailed discussions in Section \ref{sec:ana}). As the saying goes, "the rotten apple injures its neighbors", training on random examples would bring training bias from meaningless data, so it needs to be corrected with more data and results in more cost of resources and time. In addition, the inaccurate pre-training may affect the model robustness, as the PLM may fail to capture the similarity of tokens or sentences in different expressions.

%


\begin{figure*}
	\centering
	\includegraphics[width=0.88\textwidth]{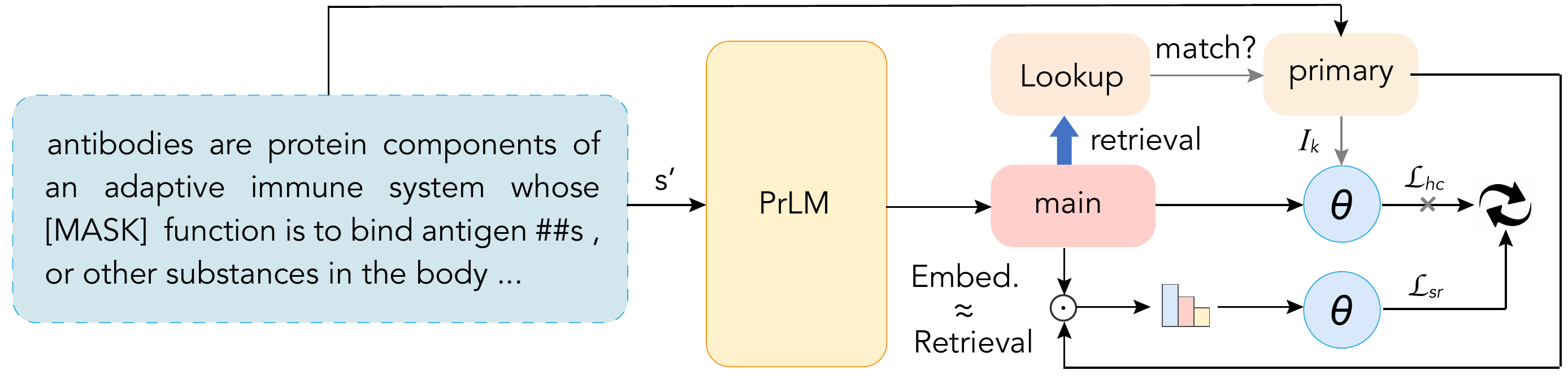}
	\vspace*{-3mm}
	\caption{Illustration of our pre-training scheme. \label{fig:framework}}
\end{figure*}

\section{Methodology}
\subsection{Masked LM}
Masked LM (MLM) is a denoising language model technique used by BERT \citep{devlin-etal-2019-bert} to take advantage of both the left and right contexts. Given a sentence $\mathbf{s} = \{w_1, w_2, \dots, w_n\}$, where a certain proportion of tokens are randomly replaced with a special mask symbol. The input is fed into the multi-head attention layer to obtain the contextual representations, which is defined as 
${H} = \textup{FFN}(\textup{MultiHead}(K,Q,V)),
$
where $K,Q,V$ are packed from the input sequence representation $\mathbf{s}$. Then, the model is trained to predict the masked token based on the context. 

Denote $\mathcal{Y}$ as the set of masked positions using the mask symbol, and {the masked tokens are represented as $w_k, k \in \mathcal{Y}$}. 
The objective of MLM is to maximize the following objective:
\begin{equation}\label{eq:mlm}
    \mathcal{L}_{mlm}(w_k, \mathbf{s}) = \mathbb{E} \Biggl(- \sum\limits_{k \in \mathcal{Y}} \log p_{\theta}(w_k\mid\mathbf{s})\Biggr).
\end{equation}

\subsection{Pre-training on True Negatives}\label{sec:techniques}
{A natural solution to encourage the language model pre-training on true negatives is to identify and counteract the false negative issue in language model pre-training. To this end, it is possible to correct or prune the harmful gradient update after detecting the false negative predictions. In detail, we investigate two enhanced pre-training objectives, including 1) hard correction (HC), which shields the gradient propagation of the false negative samples to avoid training with false negative predictions; 2) soft regularization (SR), which measures the distribution similarity between the predicted token and the original one, to smooth the tough cross-entropy by minimizing the semantic distances.} Figure \ref{fig:framework} illustrates our pre-training scheme.

\paragraph{Hard Correction}
The criteria of hard correction is to prune the gradient when the model suffers from confusion about whether the prediction is correct or not. For each prediction, we check if the predicted token $r_k$ is highly related to the ground-truth token $w_k$ based on a short lookup table $\mathcal{V}$ in which each $w_k$ is mapped to a list of synonyms $\mathcal{V}[w_k]$. The training objective is:
\begin{equation}
    \mathcal{L}_{hc} = \mathbb{E} \Biggl(-\sum\limits_{k \in \mathcal{Y}, r_k \notin \mathcal{V}[w_k]} \log p_{\theta}(w_k\mid\mathbf{s})\Biggr).
\end{equation}

In our implementation, the lookup table is built by retrieving the synonym alternatives for each word in the model vocabulary, e.g., from WordNet \citep{miller1995wordnet} or Word2Vec embedding \citep{mikolov2013distributed}. Therefore, there will be no extra computation overhead for the construction of lookup table during training and the cost of retrieving synonyms is imperceptible. For the synonym source, we use WordNet synonyms by default (Section \ref{sec:synonyms} will compare retrieving synonyms from WordNet and Word2Vec embedding). For each training iteration, if the predicted token is found in the synonym list for the gold token, then the correction is activated and the loss calculation for the $k$-th token will be neglected.\footnote{Words might be tokenized into several pieces before feeding PLMs, in which cases the correction will not be applied because those cases do not violate our criteria. We found that there are 62.51\% tokens in the model vocabulary that have synonyms found in WordNet after removing all the stopwords.}  Such a prediction will be judged as correct by HC in cross-entropy --- the correction can be applied by simply ignoring this prediction before feeding to the cross-entropy loss function. As a post-processing technique, the hard correction technique will not bring any false positives.

\paragraph{Soft Regularization}
The hard correction method above relies on external tools, which may affect the coverage of corrections due to the restricted size of the lookup table. In pursuit of more general usage, we are interested in finding a softer way to minimize the harm of false negatives. A natural way is to leverage semantic distance between the original and predicted tokens as regularization.

For $w_k$ and $r_k$, we fetch their token representations from the model's embedding layer,
denoted as $e_k$ and $e'_k$, respectively. We leverage cosine similarity as the regularization based on the intuition that the semantic distance between the prediction and gold tokens should be minimized:
\begin{equation}
    \mathcal{L}_{sr} = \frac{1}{N_m} \sum^{{N_m}}_{k=1}(1-
    \frac{e_k \cdot e'_k}{\left \| e_k \right \| \cdot \left \| e'_k \right \|}),\label{eq:sr}
\end{equation}
where $N_m$ is the number of masked tokens to predict.

SR is based on the hypothesis that the predicted tokens should have a semantic relationship with the gold ones in the same embedding space to some extent, which is supported by various existing studies \citep{bordes2013translating,zhang-zhao-2021-structural,chen2021dialogue,li2020data}.\footnote{{A possible concern of SR is that it may encourage the model to minimize the cosine similarity between unrelated tokens in early stages when the model is not well-trained. However, since MLM dominates the training, especially in the early stages (e.g., the loss from around 30 to 10 until convergence in ELECTRA-small). In contrast, the value of SR is usually 0-1. As such, SR would not harm the training in the early stages. As a regularization method, it further enhances the model performance beyond MLM when the training proceeds. }} We choose to apply SR to the embedding layer because the embedding layer is the most fundamental and stable layer {as it is far from the output layer to reflect on the returned gradients during training}. Optimizing the embedding layer would possibly lead to a more severe influence on the model training and help the model learn semantics between words better, as indicated by \cite{jiang2020smart}. Just calculating token-wise distance neglects the context of the whole sequence. In Section \ref{sec:sentence}, we will discuss the pros and cons of token-level and sentence-level SR variants.

%
    
    
  

\section{Experiments}
\subsection{Setup}\label{sec:setup}
\paragraph{Pre-training} In this part, we will introduce the model architecture, hyper-parameter setting, and corpus for pre-training our models. Our methods are applicable to general MLM-style language models. Considering the training efficiency, we employ ELECTRA small and base as our default backbone models and implement our pre-training objectives on top of them.\footnote{{How our methods work in ELECTRA is elaborated in Appendix A.}} We follow the model configurations in \cite{clark2019electra} for fair comparisons. For hyper-parameters, the batch size is 128 for the base models in our work instead of 256 as in the original setting due to limited resources. The mask ratio is 15\%. We set a maximum number of tokens as 128 for small models and 512 for base models.\footnote{For evaluation of the reading comprehension tasks, we also pre-train the variants with the length of sentences in each batch as up to 512 tokens.} The small models are pre-trained from scratch for 1000$k$ steps. To save computation, like previous studies \citep{dong2019unified}, we continue training base models for 200$k$ steps using the pre-trained weights as initialization. The learning rates for small and base models are 5e-4, and 5e-5, respectively. We use OpenWebText \citep{radford2019language} to train small models, and Wikipedia and BooksCorpus \citep{zhu2015aligning} for training base models following \cite{clark2019electra}. The baselines and our models are trained to the same steps for a fair comparison.

To verify the generality of our methods on other PLMs, we also implemented them on BERT$_{\rm{base}}$ and  BERT$_{\rm{large}}$ backbones \citep{devlin-etal-2019-bert} according to the same implementation for ELECTRA$_{\rm{base}}$. Specifically, we pre-train our methods based on BERT$_{\rm{base}}$ and BERT$_{\rm{large}}$ checkpoints for 200$k$ steps on the Wikipedia and BooksCorpus. For a fair comparison, we also train the baseline models to the same steps. Please note that it is inadequate to pursue absolute gains for large models by using single-machine NVIDIA V100 GPUs (e.g., slower convergence speed with much smaller batch sizes), compared with TPUs for training large models in public releases \cite{devlin-etal-2019-bert}. Therefore, we focus on the relevant improvements between our methods and the baselines under the same training steps.


\paragraph{Fine-tuning} For evaluation, we fine-tune the pre-trained models on GLUE (General Language Understanding Evaluation) \citep{wang-etal-2018-glue} and SQuAD v1.1 \citep{Rajpurkar2016SQuAD} to evaluate the performance of the pre-trained models. GLUE include two single-sentence tasks (CoLA \citep{warstadt2018neural}, SST-2 \citep{socher2013recursive}), three similarity and paraphrase tasks (MRPC \citep{dolan2005automatically}, STS-B \citep{cer2017semeval}, QQP \citep{chen2018quora} ), three inference tasks (MNLI \citep{nangia2017repeval},
QNLI \citep{Rajpurkar2016SQuAD}, RTE \citep{bentivogli2009fifth}. We follow ELECTRA hyper-parameters for single-task fine-tuning. We did not use any training strategies like starting from MNLI, to avoid extra distractors and focus on the fair comparison in the single-model and single-task settings.

\begin{table*}[!htb]
    \centering
    \setlength{\tabcolsep}{8.8pt}
    \setlength{\belowcaptionskip}{3pt}

    \begin{tabular}{l c c c c c c c c c c}
    \toprule
   {\textbf{Model}} & \textbf{CoLA} & \textbf{SST} & \textbf{MRPC} & \textbf{STS} & \textbf{QQP} & \textbf{MNLI} & \textbf{QNLI} & \textbf{RTE} & \textbf{Average} &$\Delta$\\
    \midrule
        BERT$_{\rm{base}}$ & 61.1 & 93.0 & 86.8 & 87.1 & 90.8  & 	84.7 & 91.4 & 67.9 & 82.9 & - \\
     BERT$^{\rm{HC}}_{\rm{base}}$ & 62.9  & 	93.2  & 	87.5  & 	87.4  & 	90.9  & 	84.9  & 	91.5  & 	69.3  & 	83.5 & $\uparrow$0.7\\
    BERT$^{\rm{SR}}_{\rm{base}}$ & 61.2  & 	93.5  & 	89.0 	 & 87.5  & 	90.9 &  	84.8  & 	91.6  & 	68.6  & 	83.4 & $\uparrow$0.6\\
   
    \midrule
    BERT$_{\rm{Large}}$ & 61.7 &	93.7 &	88.5 &	90.1 &	91.3 	& 86.7 &	92.4 &	72.9 &	84.7  & - \\
    BERT$^{\rm{HC}}_{\rm{large}}$ & 62.3 &	93.4 &	89.0 &	90.5 	& 91.5 &	87.0 &	93.0 &	73.7 &	85.1 & $\uparrow$0.4\\
    BERT$^{\rm{SR}}_{\rm{large}}$ & 62.3 &	94.2 &	89.2 &	90.1 	& 91.4 &	87.0 &	92.8 &	74.0 &	85.1  & $\uparrow$0.4 \\
    \midrule
    ELECTRA$_{\rm{small}}$ & 56.8 & 88.3 & 87.4 & 86.8 & 88.3 & 78.9 & 87.9 & 68.5 & 80.4 & -\\
     ELECTRA$^{\rm{HC}}_{\rm{small}}$ & \textbf{62.0} & 89.8 & 87.0 & 86.7 & 89.0 & 80.4 & 88.0 & 67.9 & 81.4  & $\uparrow$1.0 \\
    ELECTRA$^{\rm{SR}}_{\rm{small}}$ & 61.1 & \textbf{90.1} & \textbf{89.5} & \textbf{87.0} & \textbf{89.4} & \textbf{80.8} & \textbf{88.8} & \textbf{68.6} & \textbf{81.9} & $\uparrow$1.5 \\
   
    \midrule
    ELECTRA$_{\rm{base}}$ & 68.3 & 95.3 & 90.9 & \textbf{91.3} & 91.7 & 88.5 & 93.0 & 82.3 & 87.7 & -\\
    ELECTRA$^{\rm{HC}}_{\rm{base}}$ & \textbf{70.9} & \textbf{95.6} & \textbf{91.2} & \textbf{91.3} & \textbf{92.0} & 88.7 & \textbf{93.6} & 83.8 & \textbf{88.4}  & $\uparrow$0.7 \\
    ELECTRA$^{\rm{SR}}_{\rm{base}}$ & 70.4 & 95.4 & 90.4 & 91.2 & 91.9 & \textbf{89.1} & 93.4 & \textbf{84.8} & 88.3  & $\uparrow$0.6 \\

    
    
    \bottomrule
    \end{tabular}
        \caption{Comparisons between our proposed methods and the baseline pre-trained models on the dev set of GLUE tasks. STS is reported by Spearman correlation, CoLA is reported by Matthew's correlation, and other tasks are reported by accuracy.}
    \label{GLUE}
  \vspace*{-3mm}
\end{table*}

\begin{table}[!htb]
\centering
\vspace*{-3mm}
 \setlength{\tabcolsep}{8.8pt}
      
    \begin{tabular}{lllll}
    \toprule
    \textbf{Model}    & \textbf{EM} & $\Delta$\textbf{EM} & \textbf{F1} & $\Delta$\textbf{F1}\\
    \midrule
    ELECTRA$_{\rm{small}}$ & 75.8 & - & 83.9 & - \\
    ELECTRA$^{\rm{HC}}_{\rm{small}}$ & \textbf{77.7} & $\uparrow$1.9 & \textbf{85.6} & $\uparrow$1.7 \\
    ELECTRA$^{\rm{SR}}_{\rm{small}}$ & 76.0 & $\uparrow$0.2 & 84.2 & $\uparrow$0.3\\
    
    \midrule
    ELECTRA$_{\rm{base}}$ & 85.1 & - & 91.6 & - \\
    ELECTRA$^{\rm{HC}}_{\rm{base}}$ & \textbf{85.7} & $\uparrow$0.6 & \textbf{92.1} &$\uparrow$0.5 \\
    ELECTRA$^{\rm{SR}}_{\rm{base}}$ & 85.6 & $\uparrow$0.5  & 92.0 & $\uparrow$0.4 \\
    
    \bottomrule
  \end{tabular}
  
  \caption{Results on the SQuAD dev set. EM and F1 are short for the exact match and F1 scores \citep{Rajpurkar2016SQuAD}.}\label{exp-squad}
  
  \vspace*{-3mm}
\end{table}

\subsection{Main Results}

We evaluate the performance of our pre-training enhancement compared with the baselines in small and base sizes on GLUE and SQuAD benchmarks in Tables \ref{GLUE}-\ref{exp-squad}. From the results, we have the following observations:

1) The models with our enhanced pre-training objectives outperform the BERT and ELECTRA baselines in all the subtasks. In particular, with the same configuration and pre-training data, for both the small-size and the base-size, our methods outperform the strong ELECTRA baselines by +1.5(dev)/+1.4(test) and +0.7(dev)/+1.3(test) on average, respectively.\footnote{Test results are listed in Appendix B to save space.} The results demonstrate that our proposed methods improve the pre-training of ELECTRA substantially and disclose that mediating the training with true negatives is quite beneficial for improving language model pre-training.

2) Our methods outperform the baselines on both the base and large models,\footnote{Since larger models obtain better baseline results, we also calculate error-reduction ratio (ERR) for comparison, e.g., the ERR of BERT$^{\rm{HC}}_{\rm{base}}$ and BERT$^{\rm{HC}}_{\rm{large}}$ is 3.6\% and 2.3\%, respectively. The statistics also indicate consistent strength in varied model sizes.} which indicates that the false negative issue may be independent of the model size, and the training remains insufficient in training language models on different scales. 

3) Both SR and HC pre-training strategies help the resulting model surpass the baselines. Note that our proposed method is model-agnostic so that the convenient usability of its backbone precursor can be kept without architecture modifications. In comparison, SR is more generalizable as it does not require extra resources, while HC has the advantage of interpretation via explicit correction.

4) Our enhanced pre-training objectives show considerable performance improvements on linguistics-related tasks such as CoLA and MRPC. These tasks are about linguistic acceptability and paraphrase/semantic equivalence relationship. Besides, our methods also achieve obvious gains in tasks requiring complex semantic understanding and reasoning, such as MNLI and SQuAD, showing that they may help capture semantics to some extent.


\begin{table*}[!htb]
\centering
\setlength{\tabcolsep}{6.8pt}

{
\begin{tabular}{lcccccccc}
\toprule
\multirow{2}{*}{\textbf{Model}}
& \multicolumn{4}{c}{\textbf{\emph{AddSentenceDiverse}} (Ori.$\rightarrow$Trans.)}
& \multicolumn{4}{c}{\textbf{\emph{SwapSynWordNet}} (Ori.$\rightarrow$Trans.)}
\\
& Exact Match & $\Delta$EM & F1 Score & $\Delta$F1
& Exact Match & $\Delta$EM & F1 Score & $\Delta$F1
\\
\midrule
ELECTRA$_{\rm{small}}$ & 80.55$\rightarrow$25.60 & $\downarrow$54.95 & 85.10$\rightarrow$26.43 & $\downarrow$58.67 & 80.67$\rightarrow$74.67 & $\downarrow$6.00 & 85.38$\rightarrow$80.43 & $\downarrow$4.95 \\
ELECTRA$^{\rm{HC}}_{\rm{small}}$  & 82.59$\rightarrow$34.13 & $\downarrow$48.46 & 86.78$\rightarrow$36.60 & $\downarrow$50.18 & 82.33$\rightarrow$\textbf{79.67} & $\downarrow$\textbf{2.66} & 86.68$\rightarrow$\textbf{83.65} & $\downarrow$3.03 \\
ELECTRA$^{\rm{SR}}_{\rm{small}}$  & 78.84$\rightarrow$\textbf{37.20} & $\downarrow$\textbf{41.64} & 80.84$\rightarrow$\textbf{38.29} & $\downarrow$\textbf{42.55} & 78.67$\rightarrow$75.67 & $\downarrow$3.00 & 80.88$\rightarrow$78.51 & $\downarrow$\textbf{2.37} \\

\bottomrule
\end{tabular}
}

\caption{Robustness evaluation on the SQuAD dataset. Ori. represents the results of original dataset {for robustness evaluation} derived from the SQuAD 1.1 dev set by TextFlint \citep{wang2021textflint} while Trans. indicates the transformed one. The assessed models are the ${\rm small}$ models from Table \ref{exp-squad}.}\label{tab:robust}

\vspace*{-3mm}
\end{table*}

\begin{table*}[!htb]
    \centering
    \setlength{\tabcolsep}{8.2pt}

    {
    \begin{tabular}{l c c c c c c c c c c}
    \toprule
    \textbf{Model} & \textbf{CoLA} & \textbf{SST} & \textbf{MRPC} & \textbf{STS} & \textbf{QQP} & \textbf{MNLI} & \textbf{QNLI} & \textbf{RTE} & \textbf{Average} &$\Delta$ \\
    \midrule
    ELECTRA$_{\rm{small}}$ & 56.8 & 88.3 & 87.4 & 86.8 & 88.3 & 78.9 & 87.9 & 68.5 & 80.4 & -\\
    \midrule
    ELECTRA$^{\rm{HC}}_{\rm{WordNet}}$ & 62.0 & 89.8 & 87.0 & 86.7 & 89.0 & 80.4 & 88.0 & 67.9 & 81.4  & $\uparrow$1.0 \\
    ELECTRA$^{\rm{HC}}_{\rm{Embedding}}$ & 59.0 & 88.5 & 87.0 & 86.4 & 88.8 & 79.6 & 87.9 & 67.1 & 80.6  & $\uparrow$0.2 \\
    \midrule
    ELECTRA$^{\rm{SR}}_{\rm{Word}}$ & 61.1 & 90.1 & 89.5 & 87.0 & 89.4 & 80.8 & 88.8 & 68.6 & 81.9  & $\uparrow$1.5\\
    ELECTRA$^{\rm{SR}}_{\rm{Sent}}$ & 59.5 & 89.6 & 90.0 & 86.7 & 89.1 & 80.4 & 90.0 & 68.2 & 81.6 & $\uparrow$1.2 \\
    
    \bottomrule
    \end{tabular}
    }
    
        \caption{Comparative studies of variants on GLUE dev sets on small models. The first block shows our baseline. The second block presents the results of HC methods based on WordNet and Word2Vec embedding. The third block compares the word-level regularization and sentence-level regularization. }
    \label{tab:abl}
    
    \vspace*{-3mm}
\end{table*}

\begin{figure*}[!htb]
	\centering
	\includegraphics[width=1.0\textwidth]{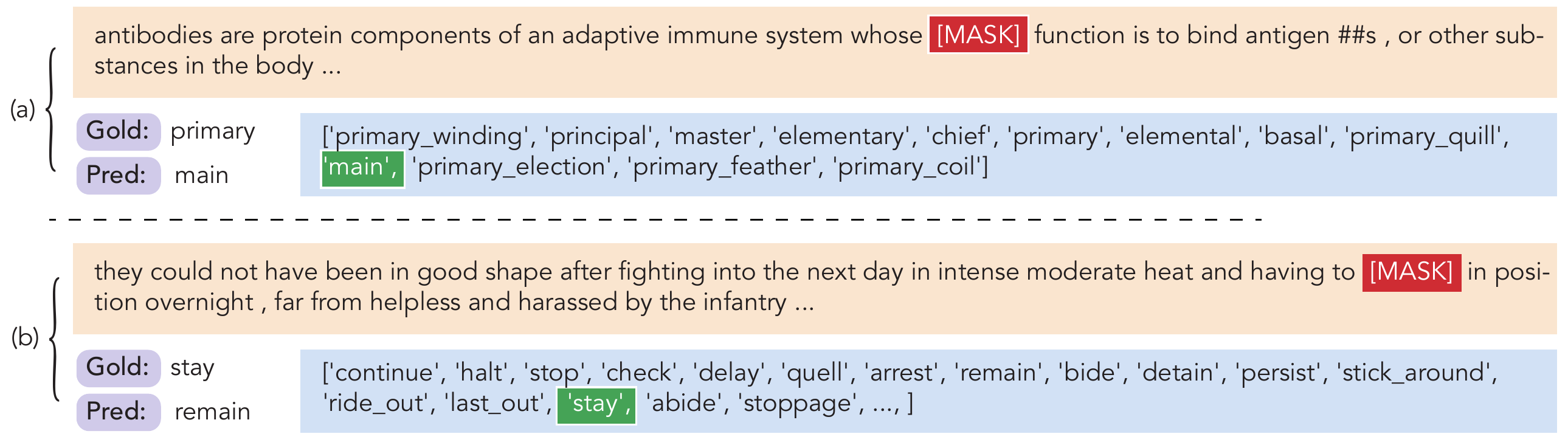}
	\caption{Interpretation of the hard correction process. The orange box contain the input sentence, the purple buttons indicate the gold and predicted tokens, and the blue box shows the WordNet synonyms for the gold token.}
	\label{fig:case}
	\setlength{\belowcaptionskip}{3pt}
	\vspace*{-3mm}
\end{figure*}

\section{Analysis}\label{sec:ana}
\paragraph{Robustness Evaluation}
Intuitively, our method would be helpful for improving the robustness of PLMs because the approaches may indicate lexical semantics and representation diversity during the correction or regularization operations. To verify the hypothesis, 
we use a robustness evaluation platform TextFlint \citep{wang2021textflint} on SQuAD, from which two standard transformation methods are adapted: 1) \textit{AddSentenceDiverse} generates distractors with altered questions and fake answers; 2) \textit{SwapSynWordNet} transforms an input by replacing its words with synonyms provided by WordNet.

Table \ref{tab:robust} shows the robustness evaluation results. We observe that both kinds of attacks induce a significant performance drop of the baseline system, by 54.95\% and 6.0\% on the EM metrics, respectively, indicating that the system is sensitive to distractors with similar meanings. In contrast, both of our models can effectively resist those attacks with less performance degradation. Specifically, the HC method works stably in the \textit{SwapSynWordNet} attack. We speculate the reason is that the hard correction strategy captures the synonym information during pre-training, which would take advantage of lexical semantics. The other variant, the soft regularization objective, achieves much better performance in the \textit{AddSentenceDiverse}. The most plausible reason might be the advantage of acquiring semantic diversity by regularizing the semantic distance in the SR objective. 


\paragraph{Lookup Table from WordNet vs. Word2Vec}\label{sec:synonyms}
For the hard correction approach, the candidate synonyms for detecting false negative predictions can be derived from WordNet \citep{miller1995wordnet} or Word2Vec embedding space \citep{mikolov2013distributed} as described in Section \ref{sec:techniques}.\footnote{We use the public \texttt{GloVe.6B.50d} vectors for embedding retrieval. Since the embedding method returns a ranked list by calculating the similarity score with the whole vocabulary, we only take the top 10 most similar words for each retrieval.} To verify the impact of different sources, we compare the results as shown in the second block of Table \ref{tab:abl}. We see that ELECTRA$^{\rm{HC}}_{\rm{WordNet}}$ outperforms ELECTRA$^{\rm{HC}}_{\rm{Embedding}}$ by a large margin. The most plausible reason would be that the retrieved list of synonyms from ELECTRA$^{\rm{HC}}_{\rm{WordNet}}$ would have higher quality than that from ELECTRA$^{\rm{HC}}_{\rm{Embedding}}$. Although the embedding-based method may benefit from semantic matching, but would also bring noise as it is hard to set the threshold to ensure the top-ranked words are accurate synonyms. Therefore, ELECTRA$^{\rm{HC}}_{\rm{WordNet}}$ turns out to be better suitable for our task.\footnote{Since the HC method relies on synonym retrieval from external sources, it is possible to bring false positive corrections theoretically. However, we seldom detect such cases in our preliminary experiments, so we leave the open question for interested readers to avoid deviating from the focus of this paper. }

\paragraph{From Word-level to Sentence-level Regularization}\label{sec:sentence}
The soft regularization approach measures the semantic distance between the predicted one and the ground truth, which may neglect the sentence-level context. We are interested in whether measuring the sentence-level similarity would achieve better results. To verify the hypothesis, we fill the masked sentence  $\mathbf{s}$ with the predicted tokens $r_k$ to have the predicted sentence $s_p$. Then, $s_p$ and $s$ are fed to the Transformer encoder to have the contextualized representation $H_p$ and $H_s$, respectively. To guide the probability distribution of model predictions $H_p$ to match the expected probability distribution $H_s$, we adopt Kullback–Leibler (KL) divergence:
$\mathcal{L}_{kl} = \textup{KL}(H_p \parallel H_s),\label{eq:kl}$
where $\mathcal{L}_{kl}$ is applied as the degree of sentence-level semantic mismatch. {In detail, we first apply softmax on the two hidden representations to obtain two distributions, and then the KL divergence is calculated between those two distributions.} The loss function is then written as:
$
\mathcal{L'} = \mathcal{L}_{dlm} + \mathcal{L}_{kl}. 
$
For clarity, we denote the original ELECTRA$^{\rm{SR}}_{\rm{small}}$ method described in Eq. \ref{eq:sr} as ELECTRA$^{\rm{SR}}_{\rm{Word}}$ and the sentence-level variant as ELECTRA$^{\rm{SR}}_{\rm{Sent}}$. 

The comparative results are reported in the third block of Table \ref{tab:abl}, which indicates that using sentence-level regularization (ELECTRA$^{\rm{SR}}_{\rm{Sent}}$) also outperforms the baseline and nearly reaches the performance of word-level one (ELECTRA$^{\rm{SR}}_{\rm{Word}}$) on average, with slightly better results on MRPC and MNLI. Although ELECTRA$^{\rm{SR}}_{\rm{Sent}}$ still keeps the same parameter size with baseline, it leads to more computation resources because it requires the extra calculation for the predicted sequence $H_p$. Therefore, considering the balance between effectiveness and efficiency, ELECTRA$^{\rm{SR}}_{\rm{Word}}$ can serve as the first preferred choice for practical applications, and ELECTRA$^{\rm{SR}}_{\rm{Sent}}$ can be employed when computation resources are sufficient.\footnote{{Besides the methods we discussed in this work, there are alternative ways to achieve the regularization effects, e.g., using a softmax temperature with the standard loss.}}


\paragraph{Case Studies} To interpret how our method works, we randomly select some hard correction examples as shown in Figure \ref{fig:case} by taking the ELECTRA$_{\rm{small}}$ as the baseline model. We find that the baseline model produces reasonable predictions such as \textit{main} and \textit{remain}, as opposed to the golds ones, \textit{primary} and \textit{stay}. Those predictions will be determined as wrong and possibly harm pre-training. Fortunately, such cases can be easily solved by our proposed method. Though the synonym list may contain irrelevant words, our correction will not bring false positives because it only cares about whether the predicted word is in the shortlist or not. 
Being a detected synonym is a sufficient condition, though it is not a necessary condition as those predictions make up the subset of false negatives. Therefore, inappropriate options in the list would not bring side effects.



\paragraph{Pre-training Efficiency} The training cost is nearly the same as the baseline, {i.e., The training time for the BERT-base baseline/HC/SR are 100/102/101 hours for 200K steps using a single machine with NVIDIA V100 GPUs.}. The possible cost includes lookup table construction and retrieval. The lookup table is built offline by retrieving synonyms for each token from model vocabulary, which took 5 seconds. The cost of retrieving synonyms is imperceptible.

\section{Related Work}
Self-supervised learning is one of the major topics in training pre-trained models \citep{peters2018deep,radford2018improving,devlin-etal-2019-bert,zhu2022leveraging}, which decides how the model captures knowledge from large-scale unlabeled data. Recent studies have investigated denoising patterns \citep{raffel2020exploring,lewis2020bart}, MLM alternatives \citep{yang2019xlnet}, and auxiliary objectives \citep{lan2019albert,joshi-etal-2020-spanbert} to improve the power of pre-training. However, studies show that the current models still suffer from under-fitting issues, and it remains challenging to find efficient training strategies \citep{rogers2020primer}. 

\paragraph{Denoising Patterns} MLM has been widely used as the major objective for pre-training \citep{devlin-etal-2019-bert,lan2019albert,clark2019electra,song2020mpnet}, in which the fundamental part is how to construct high-quality masked examples \citep{raffel2020exploring}. The current studies commonly define specific patterns for mask corruption. For example, some are motivated from the language modeling units, such as subword masking \citep{devlin-etal-2019-bert}, span masking \citep{joshi-etal-2020-spanbert}, and $n$-gram masking \citep{levine2021pmimasking,li-zhao-2021-pre}. Some employ edit operations like insertion, deletion, replacement, and retrieval \citep{lewis2020bart,guu2020realm}. Others seek for external knowledge annotations, such as named entities \citep{sun2019ernie}, semantics \citep{zhou2020limit,zhang2020semantics}, and syntax \citep{zhang2020sg,xu-etal-2021-syntax}. To provide more diversity of mask tokens, RoBERTa applied dynamic masks in different training iterations \citep{liu2019roberta}. These prior studies either employ pre-defined mask construction patterns or improve the diversity of mask tokens to help capture knowledge from pre-training. 

\paragraph{MLM Alternatives} To alleviate the task mismatch between the pre-training and the fine-tuning tasks, XLNet \citep{yang2019xlnet} proposed an autoregressive objective for language modeling through token permutation, which further adopts a more complex model architecture. Instead of corrupting sentences with the mask symbol that never appears in the fine-tuning stage, MacBERT \citep{cui2020revisiting} proposes to use similar words for the masking purpose. \citet{yamaguchi2021frustratingly} also investigates simple pre-training objectives based on token-level classification tasks as replacements of MLM, which are often computationally cheaper and result in comparable performance to MLM. In addition, training  sequence-to-sequence (Seq2Seq) language models has also aroused continuous interests \citep{dong2019unified,lewis2020bart,raffel2020exploring}.

\paragraph{Auxiliary Objectives} Another research line is auxiliary objectives in conjunction with MLM, such as next sentence prediction \citep{devlin-etal-2019-bert}, span-boundary objective \citep{joshi-etal-2020-spanbert}, and sentence-order prediction \citep{lan2019albert}. Such line of research emerges as hot topics, especially in domain-specific pre-training, such as dialogue-oriented language models, which involve diverse kinds of interaction entailed in utterances \citep{zhang2020dialogpt,wu2020tod,zhang-zhao-2021-structural}. 

As the major difference from the existing studies, our work devotes itself to mediating misconceived negatives as the essential drawback of MLM during the MLE estimation and aiming to guide language models to learn from true negatives through our newly proposed regularization and correction methods. Besides the heuristic pre-trained patterns like masking strategies during data construction, we stress that there are potential post-processing strategies to guide the MLM training: correction and pruning. Those strategies are considered to deal with the false negative issue during MLM training, where the model would yield reasonable predictions but discriminated as wrong predictions because such predictions do not match the single gold token for each training case. For example, many tokens are reasonable but written in different forms or are the synonyms of the expected gold token. We could directly drop the uncertain predictions or correct the training with soft regularization. Promoting our view to sentence level, the similarity between the predicted sentence and the original sentence can also be taken into account to measure the sentence-level confidence that indicates how hard the task is, which would be beneficial to provide more fine-grained signals and thus improve the training quality. Based on the rationales above, we are motivated to design the corresponding correction and regularization techniques to mediate misconceived negatives.



\section{Conclusions}
The work identifies the false negative issue in language model pre-training and proposes methods to counteract it.
Though discriminative PLMs may quite straightforwardly suffer from the false negative issue according to our exploration in this work, it has been completely ignored for a long time, and it is a bit surprising that maybe this work is the first one that formally considers such a big pre-training leak. To counteract the intrinsic and critical issue, we investigate pre-training objectives to correct or prune the harmful gradient update after detecting the false negative predictions. Experimental results verify the superiority of our pre-training enhancement. Robustness evaluation shows that our methods can help the resulting PLM effectively resist various attacks while existing common PLMs suffer from significant performance degradation. To our best knowledge, this is also the first work to consider model effectiveness and robustness of language model pre-training simultaneously. Our work indicates that mediating false negatives is so important that counter-false-negative pre-training can synchronously improve the effectiveness and robustness of PLMs.


\bibliography{aaai23}

\clearpage
\appendix
\section*{{Appendix}}

\begin{table*}[htb]
    \centering
    \setlength{\tabcolsep}{3.8pt}
    \setlength{\belowcaptionskip}{3pt}
    \caption{{Comparisons with public methods on GLUE test sets. The public results in the first column are from BERT \citep{devlin-etal-2019-bert}, SpanBERT \citep{joshi-etal-2020-spanbert}, and ELECTRA \citep{clark2019electra}. The second columns present our implementations.} }
    \label{tab:pub}
     
    \begin{tabular}{l c c c c c c c c c c c }
    \toprule
    \textbf{Model} & \textbf{Params} & \textbf{CoLA} & \textbf{SST} & \textbf{MRPC} & \textbf{STS} & \textbf{QQP} & \textbf{MNLI} & \textbf{QNLI} & \textbf{RTE} & \textbf{Average} &$\Delta$ \\
    \midrule        
    BERT$_{\rm{base}}$ &  110M & 52.1 & 93.5 & 84.8 & 85.8 & 89.2 & 84.6 & 90.5 & 66.4 & 80.9 & -  \\
    BERT$_{\rm{large}}$ &  335M & 60.5 & 94.9 & 85.4 & 86.5 & 89.3 & 86.7 & 92.7 & 70.1 & 83.3 & -\\
    SpanBERT$_{\rm{large}}$ &  335M & 64.3 & 94.8 & 87.9 & 89.9 & 89.5 & 87.7 & 94.3 & 79.0 & 85.9 & - \\
    ELECTRA$_{\rm{small}}$ &  14M &  54.6 &  89.1 &  83.7 &  80.3 &  88.0 &  79.7 &  87.7 &  60.8 &  78.0& - \\
    ELECTRA$_{\rm{base}}$ & 110M & 59.7 & 93.4 & 86.7 & 87.7 & 89.1 & 85.8 & 92.7 & 73.1 & 83.5 & -\\
    \midrule
    ELECTRA$_{\rm{small}}$ &  14M & 53.6 & 88.9 & 83.9 & 80.7 & 87.1 & 80.1 & 87.9 & 63.1 & 78.2 & - \\
     ELECTRA$^{\rm{HC}}_{\rm{small}}$ &  14M  & 55.3 &  90.3 &  84.1 &  {82.0} &  87.2 &  {80.6} & {88.4} &  64.3 &  79.0  &  $\uparrow$0.8 \\
    ELECTRA$^{\rm{SR}}_{\rm{small}}$ &  14M  &  {58.3} &  {90.6} & {85.4} &  81.4 &  {87.9} &  {80.6} &  88.0 &  	{64.3} &  {79.6} & $\uparrow$1.4 \\
   ELECTRA$_{\rm{base}}$ & 110M & 62.4 & 95.3 & 87.3 & 88.8 & 89.0 & 87.6 & 93.1 & 77.6 & 85.1   & -   \\
   ELECTRA$^{\rm{HC}}_{\rm{base}}$ & 110M  & \textbf{67.5} & \textbf{95.8} & \textbf{88.6} & 89.9 & 89.7 & 89.0 & \textbf{93.6} & \textbf{79.1} & \textbf{86.7}  & $\uparrow$1.6 \\
    ELECTRA$^{\rm{SR}}_{\rm{base}}$  & 110M & 65.7 & 95.7 & 88.3 & \textbf{90.0} & \textbf{89.9} & \textbf{89.1} & \textbf{93.6} & 78.8 & 	86.4  & $\uparrow$1.3 \\
    
    \bottomrule
    \end{tabular}
    
\end{table*}

\section{{How HC and SR work in ELECTRA}}\label{app:electra}
{The two proposed methods, HC and SR, affect both the generator and discriminator of ELECTRA. If HC finds a position whose prediction is the synonym of the original masked word, the generator will ignore the prediction in that position. Since the generator will be fed with the gold labels (original token ids) for each position, we can easily convert the labels of the concerned position into -100 (the index to be ignored in CrossEntropy) in the PyTorch implementation. Similarly, for the discriminator, the position will also be considered as the original word instead of the replaced one. For SR, since the generator and discriminator share the embeddings, jointly training with SR can contribute to the overall loss and the gradient will be passed thorough back to the Embedding layer to guide the training of both modules.}

\section{{Test Results on the GLUE datasets}}
Table \ref{tab:pub} shows the comparison with public models on the GLUE test set.\footnote{Due to the limited frequency of online submissions, we only submitted ELECTRA models for online tests.} Compared with the public methods, our model not only far exceeds the performance of others under the same model scale but also outperforms some large public models by our base settings with much fewer parameters. {There are recent advances beyond ELECTRA models \cite{meng2021coco,hao2021learning,meng2021pretraining}, which are not included in comparisons because the training settings vary quite much, e.g., training data, batch sizes, model architectures, etc.}

\end{document}